\documentclass[10pt,twocolumn,letterpaper]{article}

\usepackage[pagenumbers]{wacv} 
\definecolor{wacvblue}{rgb}{0.21,0.49,0.74}
\usepackage{xurl} 

\usepackage[pagebackref,breaklinks,colorlinks]{hyperref}
\hypersetup{
  allcolors = wacvblue
}

\usepackage{booktabs} 
\usepackage{placeins}  
\usepackage{multirow}
\usepackage{caption}
\usepackage{microtype}  

\usepackage{amssymb}  
\usepackage{pifont}
\usepackage{amsmath}

\usepackage{algorithm}
\usepackage{algorithmic}
\usepackage{hhline}

\usepackage{tabularx}

\newcommand{\tr}[1]{\textcolor{red}{#1}}

\usepackage{svg}
\usepackage[utf8]{inputenc}
\usepackage{kotex} 

\def\wacvPaperID{1981} 
\def\confName{WACV}
\def\confYear{2026}

\title{RS-Net: Context-Aware Relation Scoring for Dynamic Scene Graph Generation}

\author{
Hae-Won Jo and Yeong-Jun Cho\thanks{Corresponding author: \texttt{yj.cho@jnu.ac.kr}}\\
Department of Artificial Intelligence Convergence\\
Chonnam National University, Gwangju 61186, South Korea\\
{\tt\small \{haewon\_jo, yj.cho\}@jnu.ac.kr}
}

\begin{document}

\maketitle

\begin{abstract}

Dynamic Scene Graph Generation (DSGG) models how object relations evolve over time in videos. However, existing methods are trained only on annotated object pairs and lack guidance for non-related pairs, making it difficult to identify meaningful relations during inference.
In this paper, we propose Relation Scoring Network (RS-Net), a modular framework that scores the contextual importance of object pairs using both spatial interactions and long-range temporal context. RS-Net consists of a spatial context encoder with learnable context tokens and a temporal encoder that aggregates video-level information. The resulting relation scores are integrated into a unified triplet scoring mechanism to enhance relation prediction.
RS-Net can be easily integrated into existing DSGG models without architectural changes. Experiments on the \texttt{Action} \texttt{Genome} dataset show that RS-Net consistently improves both Recall and Precision across diverse baselines, with notable gains in mean Recall, highlighting its ability to address the long-tailed distribution of relations. Despite the increased number of parameters, RS-Net maintains competitive efficiency, achieving superior performance over state-of-the-art methods.
Our code is available at \href{URL}{https://}
\end{abstract}




\section{Introduction}


Scene Graph Generation (SGG) aims to represent an image as a structured graph, where nodes correspond to detected objects and edges represent their semantic relationships. 
By providing an interpretable and compact abstraction of visual content, scene graphs have been widely used in various vision-language tasks such as image captioning, visual question answering, and image retrieval. 
To extend this structured understanding to the video domain, Dynamic Scene Graph Generation (DSGG) has emerged, focusing on modeling how objects and their interactions evolve over time across frames.
Recent studies on DSGG have focused on capturing temporal object interactions through local context modeling~\cite{cong2021spatial} and architectural innovations such as spatio-temporal transformers and hierarchical structures~\cite{feng2023exploiting, nguyen2024hig}. 
These approaches have improved the temporal coherence and structural quality of scene graphs in dynamic environments.

\begin{figure}[t]
    \centering
    \includegraphics[width=1\linewidth]{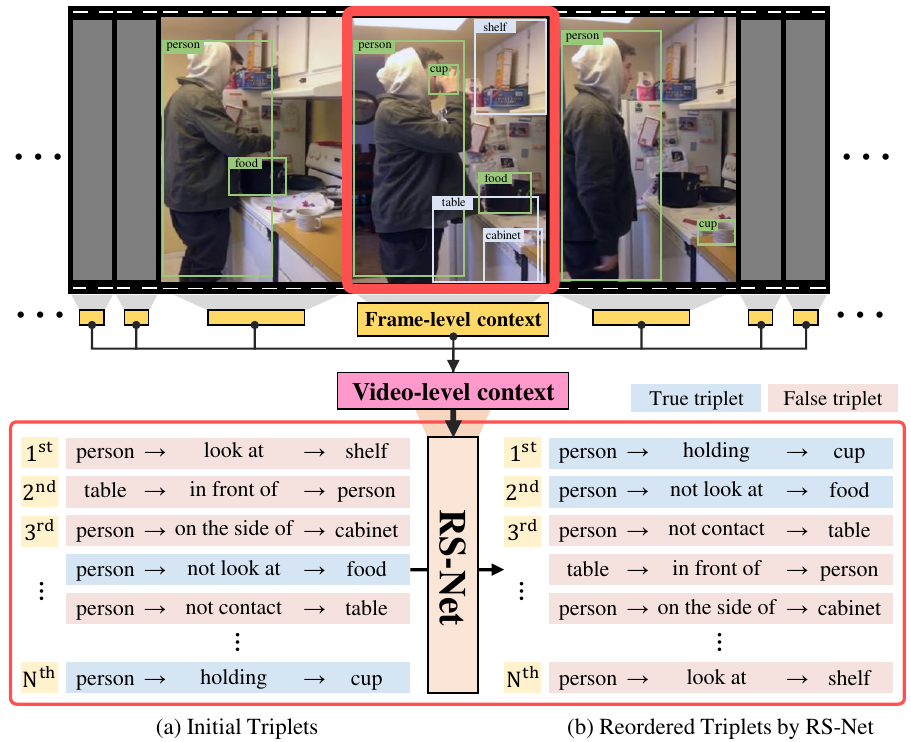}
    \caption{
        Example of our proposed relation scoring method. 
        (a) Initial triplet predictions from existing DSGG methods. (b) Triplets reordered by our proposed relation scoring approach, where contextually important relations are ranked higher based on video-level semantic relevance.
    }
    \label{fig:figure_1}
\end{figure}


Despite recent progress, existing DSGG methods still face two major limitations.
First, they are trained only on object pairs with annotated relations, while pairs without annotations are ignored. 
This gap between limited supervision and exhaustive predicate prediction at inference time creates a distributional mismatch. 
As a result, the model often fails to distinguish meaningful relations from irrelevant ones in complex dynamic scenes.
Second, many approaches~\cite{feng2023exploiting,wang2024oed} process only short temporal segments (typically 2--3 consecutive frames), which limits their ability to capture long-range dependencies and global context across the entire video.



To address these limitations, we propose RS-Net, a relation scoring network designed to improve relational reasoning in dynamic scenes.
RS-Net introduces a scoring mechanism that evaluates the contextual importance of each object pair, helping the model distinguish meaningful relations from irrelevant ones (Fig.~\ref{fig:figure_1}).
In addition, it incorporates a temporal context encoder that captures long-range dependencies across the entire video, rather than relying on short sliding-window segments.
This design enables RS-Net to better model global context and temporal dynamics, allowing it to evaluate the importance of each relation.

RS-Net is composed of three key modules: a spatial context encoder, a temporal context encoder, and a relation scoring decoder. The spatial context encoder extracts intra-frame object interactions using a Transformer backbone and a learnable spatial context token, which aggregates semantic cues from the entire frame. The temporal context encoder captures long-range dependencies across frames by attending to these spatial tokens over time and integrating them into a temporal context token that summarizes video-level context. Finally, the relation scoring decoder computes a relation score for each object pair by combining local interaction features with global contextual cues from both encoders.


RS-Net can be easily integrated into existing DSGG frameworks without structural changes.
It enhances relation prediction by incorporating contextual information and relation scores into the existing triplet structure.
Experimental results on the \texttt{Action} \texttt{Genome} dataset~\cite{ji2020action} demonstrate that RS-Net not only consistently improves performance across various DSGG baselines, but also achieves superior results in both Recall and Precision.
When applied to STTran~\cite{cong2021spatial}, RS-Net achieves up to +3.2 R@10 improvement in Recall, along with consistent gains in Precision (+2.7 P@10).
These results confirm that RS-Net enhances relational reasoning and improves the overall quality of the generated scene graphs.

The main contributions of this work are as follows:
\begin{itemize}
\item We propose RS-Net, a relation scoring network that can be seamlessly integrated into existing DSGG frameworks without structural modifications.
\item We introduce a context-aware scoring mechanism that incorporates both spatial and temporal context tokens to explicitly evaluate the importance of each object pair.
\item We demonstrate that RS-Net consistently improves Recall, Precision, and mean Recall across multiple DSGG baselines on the \texttt{Action} \texttt{Genome} dataset, while maintaining competitive efficiency.
\end{itemize}
To the best of our knowledge, RS-Net is the first attempt to perform relation importance scoring in dynamic scene graph generation by leveraging both spatial and long-range temporal context.


\begin{figure*}[t]  
    \centering
    \includegraphics[width=\linewidth]{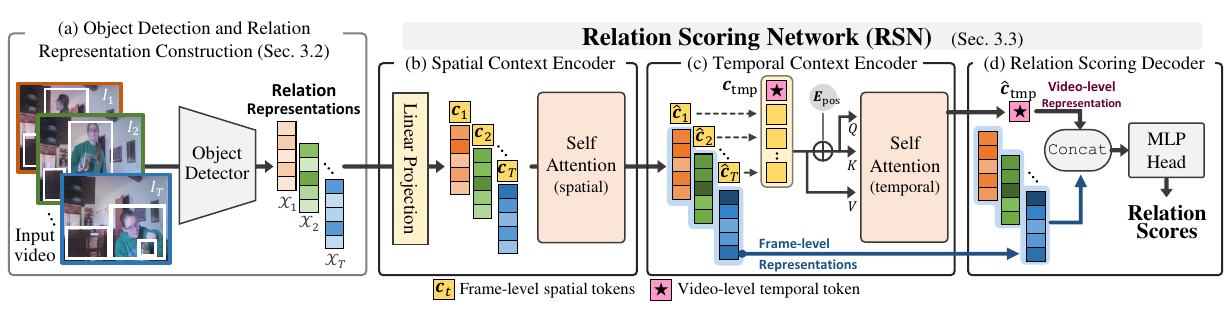}
    \caption{
        Overview of the proposed RS-Net.
        The framework consists of four main components: (a) object detection and relation representation construction, 
        (b) spatial context encoder, (c) temporal context encoder, and (d) relation scoring decoder.
        RS-Net is trained to distinguish semantically meaningful relations from irrelevant ones by incorporating both spatial and temporal cues.
        The resulting scores are used to guide predicate classification and triplet score computation during scene graph generation.
    }
    \label{fig:main_figure}
\end{figure*}


\section{Related Works}
\subsection{Scene Graph Generation.}

Scene Graph Generation~(SGG) aims to detect objects in an image and infer pairwise relationships to construct a structured representation of the visual scene.
\cite{xu2017scene} is the first work to model object relationships through scene graphs.
They proposed an end-to-end approach to model objects and their relationships, enabling context-aware relation prediction.
Building on this, subsequent SGG research has evolved toward effectively modeling complex inter-dependencies between objects and relations.
For example, Neural Motifs~\cite{zellers2018neural} captured contextual dependencies by leveraging statistical regularities in object and relation distributions, while the work~\cite{yang2018graph} modeled structural interactions using graph convolutions.

The emergence of Transformers marked a turning point in the SGG field. 
SGTR~\cite{li2022sgtr} modeled relations between entities and predicates more explicitly via bipartite graph construction, while RelTR~\cite{cong2023reltr} enabled efficient triplet generation using coupled queries in a set prediction framework. 
EGTR~\cite{im2024egtr} further achieved relation extraction by leveraging the inherent relational cues encoded in self-attention.





\subsection{Dynamic Scene Graph Generation.}
Dynamic Scene Graph Generation (DSGG) extends SGG to the video domain, aiming to capture the temporal evolution of object relationships for dynamic scene understanding beyond static image analysis.
Early approaches extended static SGG methods~\cite{lu2016visual, zellers2018neural, tang2019learning} to videos, but struggled to fully exploit the temporal and contextual information inherent in video sequences.
Since STTran~\cite{cong2021spatial} introduced spatial temporal transformers to model object interactions across frames, DSGG research has progressed along two main directions: temporal modeling refinement and architectural enhancement.

First, to refine temporal modeling, several methods have incorporated prior knowledge and structured representations to better capture spatio-temporal relations. 
For example, TRACE~\cite{teng2021target} used hierarchical relation trees and target-adaptive aggregation, while STKET~\cite{pu2023spatial} leveraged spatial and temporal priors through knowledge-embedded attention.
Second, from the architectural perspective, models have aimed to improve scene graph generation through structural innovations. DSGDetr~\cite{feng2023exploiting} enhanced temporal consistency using object tracking, and HIG~\cite{nguyen2024hig} employed hierarchical modules to model complex interactions. Furthermore, TEMPURA~\cite{nag2023unbiased}, TD2-Net~\cite{lin2024td2}, and STABILE~\cite{zhuang2025spatial} introduced techniques to address bias and stabilize training.

Despite recent progress, existing DSGG methods still have two major limitations.
First, they are trained only on object pairs with annotated relations, but during inference, they must predict relations for all detected object pairs, which makes it difficult to generalize to unseen combinations.
Second, most approaches~\cite{li2022dynamic, zhang2023end} analyze only short temporal segments (e.g., 2--3 frames), limiting their ability to model long-term interactions across the entire video.

\section{Proposed Methods}
\subsection{Motivation and Method Overview}
Dynamic Scene Graph Generation (DSGG) aims to construct a sequence of temporally evolving scene graphs ${G_t = (V_t, E_t)}_{t=1}^T$ from a video sequence $\{I_t\}^{T}_{t=1}$.
Here, $I_t$ denotes the video frame at time $t$, $V_t$ is the set of detected objects (nodes), and $E_t$ is the set of semantic relations (edges) between these objects.
A major challenge in DSGG is to distinguish contextually meaningful relations from irrelevant ones among a large number of detected objects in dynamic scenes.
Most existing methods~\cite{cong2021spatial, feng2023exploiting, pu2023spatial} rely solely on annotated object pairs with relations during training. 
However, during inference, they must predict predicates for all possible object pairs detected in each frame, regardless of whether those pairs were supervised during training.
This gap between limited training supervision and exhaustive inference introduces a distributional mismatch that limits the model’s capacity for relational reasoning in dynamic scenes.

To address this issue, we propose a Relation Scoring Network (RS-Net), as illustrated in Fig.~\ref{fig:main_figure}.
Unlike prior methods that rely solely on annotated positive relations, RS-Net is trained to learn from both semantically relevant (positive) and irrelevant (negative) object pairs.
This enables the model to more effectively distinguish meaningful relations from trivial ones across diverse visual contexts.
To this end, RS-Net employs a Transformer-based architecture to jointly model spatial and temporal cues, allowing it to assess the contextual relevance of object relations in dynamic scenes.
RS-Net integrates seamlessly into existing DSGG architectures and enhances scene graph generation by prioritizing contextually important relations.

\subsection{Object Detection and \\ \quad\quad \ Relation Representation Construction}
We utilize an object detector to generate $N_t$ object proposals from the $t$-th video frame $I_t$.
For each detected object $i$, we extract a visual feature vector $\mathbf{v}^i_t$, a bounding box $\mathbf{b}^i_t$, and a category probability distribution $\mathbf{d}^i_t$.
For each object pair $(i, j)$ in frame $t$, we construct a set of relation representations, denoted as
$\mathcal{X}_t = \{ \mathbf{x}^1_t, \mathbf{x}^2_t, \ldots, \mathbf{x}^{K(t)}_t \}$,
where $K(t)$ denotes the total number of relations extracted from frame $t$.

As in~\cite{cong2021spatial}, the relation representation $\mathbf{x}^k_t$ is constructed by concatenating the following components:
\begin{equation}
\mathbf{x}^k_t =
\left[\bar{\mathbf{v}}^i_t, \bar{\mathbf{v}}^j_t, \bar{\mathbf{u}}^{ij}_{t}, \bar{\mathbf{d}}^i_t, \bar{\mathbf{d}}^j_t \right],
\label{eq:relation_vector}
\end{equation}
where $\bar{\mathbf{v}}^i_t$ and $\bar{\mathbf{v}}^j_t$ are the linearly projected visual features of objects $i$ and $j$, respectively.
$\bar{\mathbf{d}}^i_t$ and $\bar{\mathbf{d}}^j_t$ are their semantic category embeddings, also obtained via linear projection from the original category distributions $\mathbf{d}^i_t$ and $\mathbf{d}^j_t$.
$\bar{\mathbf{u}}^{ij}_t$ denotes the union region feature, extracted from the union bounding box that spatially covers both $\mathbf{b}^i_t$ and $\mathbf{b}^j_t$, capturing their spatial interaction.

\subsection{Relation Scoring Network~(RS-Net)}
\label{sec:sce}
RS-Net aims to identify and prioritize contextually meaningful object relations by learning from both positive and negative object pairs.
To this end, it comprises three key components: 1) spatial context encoder, 2) temporal context encoder, and 3) relation scoring decoder.

\paragraph{Spatial Context Encoder.}

The meaning of a relation between two objects can depend on other relations in the same frame.
To capture these spatial contextual interactions among relations, we apply a transformer-based self-attention over all relation representations within the same frame as illustrated in Fig.~\ref{fig:main_figure}(b).
This allows the model to learn both the dependencies between relations and the contextual importance of each individual relation in the frame.

The relation representations $\mathbf{x}^k_t$ are linearly projected to a dimension $d_m$, which standardizes the input size and helps align relation features in a shared embedding space.
In addition, we introduce a learnable spatial context token $\mathbf{c}_t \in \mathbb{R}^{d_m}$ to summarize global spatial information of each frame, inspired by the class token mechanism in ViT~\cite{dosovitskiy2020image}.
We construct the spatial input representation for frame $t$ by concatenating the context token and the projected relation features as follows:
\begin{equation}
\mathbf{Z}_t^{\mathrm{Spa}} = \left[ \mathbf{c}_t,\, \mathbf{x}^1_t,\, \mathbf{x}^2_t,\, \ldots,\, \mathbf{x}^{K(t)}_t \right].
\end{equation}
This representation is then passed through a self-attention network to produce $\hat{\mathbf{Z}}_t^{\mathrm{Spa}} = \mathrm{Attn}(\mathbf{Z}_t^{\mathrm{Spa}})$,
where the first token $\hat{\mathbf{c}}_t$ of $\hat{\mathbf{Z}}_t^{\mathrm{Spa}}$ summarizes the frame-level spatial context,
and each $\hat{\mathbf{x}}_t^k$ in $\hat{\mathbf{Z}}_t^{\mathrm{Spa}}$ encodes relation-specific features enriched with spatial context.

\paragraph{Temporal Context Encoder.}
To capture temporal dependencies across the video frames, we encode the sequence of frame-level spatial context tokens $\hat{\mathbf{c}}_t$ using a Transformer-based temporal encoder as shown in Fig.~\ref{fig:main_figure}(c).
This allows the network to reason about the relative importance of each frame in the global context.
To achieve this, we construct a temporal representation by inserting a learnable temporal context token $\mathbf{c}_{\mathrm{tmp}} \in \mathbb{R}^{d_m}$ to the ordered frame-level context tokens as follows:
\begin{equation}
\mathbf{Z}^{\mathrm{Temp}} = \left[ \mathbf{c}_{\mathrm{tmp}},\, \hat{\mathbf{c}}_1,\, \hat{\mathbf{c}}_2,\,\ldots,\, \hat{\mathbf{c}}_T \right].
\end{equation}

To capture temporal order more effectively, we incorporate a learnable positional embedding $\mathbf{E}_{\mathrm{pos}} \in \mathbb{R}^{(T+1) \times d_m}$ into the self-attention computation. 
Specifically, the embedding is added to the Query and Key only, as follows:
\begin{equation}
\small
    \hat{\mathbf{Z}}^{\mathrm{Temp}} = \mathrm{Attn}(Q = K = \mathbf{Z}^{\mathrm{Temp}} + \mathbf{E}_{\mathrm{pos}},\ V = \mathbf{Z}^{\mathrm{Temp}}).
\end{equation}
This design encourages the model to attend to frame-level relations with awareness of temporal order, without modifying the value vectors.
Then, the self-attention module captures dependencies across frames and integrates temporal context into the sequence.

As a result, the output token $\hat{\mathbf{c}}_{\mathrm{tmp}}$ serves as a video-level representation that summarizes the entire sequence with awareness of inter-frame relations.
Unlike prior methods~\cite{lin2024td2,peddi2025towards} that process short temporal segments using a sliding window (typically 2--3 frames), our temporal context encoder captures long-term relational patterns and global contextual dependencies across the entire video.

\paragraph{Relation Scoring Decoder.}
To assess the contextual importance of each relation, we introduce a Relation Scoring Decoder consisting of a multi-layer perceptron (MLP) followed by a softmax layer as shown in Fig.~\ref{fig:main_figure}(d).
For each relation $\hat{\mathbf{x}}_t^k$, we concatenate it with the video-level temporal context token $\hat{\mathbf{c}}_{\mathrm{tmp}}$ and feed them into the decoder as follows:
\begin{equation}
\small
\mathbf{p}_t^k=\left[{p}_{t,0}^k,\ {p}_{t,1}^k\right] = \mathrm{softmax}\left(\mathrm{MLP}\left(\left[\hat{\mathbf{x}}_t^k,  \hat{\mathbf{c}}_{\mathrm{tmp}}\right]\right)\right) \in \mathbb{R}^{2}.
\end{equation}
Here, $\mathbf{p}_t^k$ denotes a 2-dimensional probability distribution, referred to as the relation score, which indicates whether the relation $\hat{\mathbf{x}}_t^k$ is contextually meaningful (positive) or irrelevant (negative) within the video.

The model is trained to output $\mathbf{p}_t^k = \left[1, 0\right]$ for positive relations and $\mathbf{p}_t^k = \left[0, 1\right]$ for negative ones.  
Our design produces a 2-dimensional probability distribution rather than a scalar sigmoid output, enabling the model to represent mutually exclusive relation types.  
This formulation allows stable training with the multi-label margin loss described in Sec.~\ref{sec:learning_RS-Net} and provides a clearer probabilistic interpretation.  
During inference, the predicted relation scores are used to identify meaningful subject–relation–object triplets in dynamic scene graph generation.  
This enables the model to consider both the spatial characteristics of each relation and the overall temporal context of the video.

\begin{figure}[t]
    \centering
    \includegraphics[width=1\linewidth]{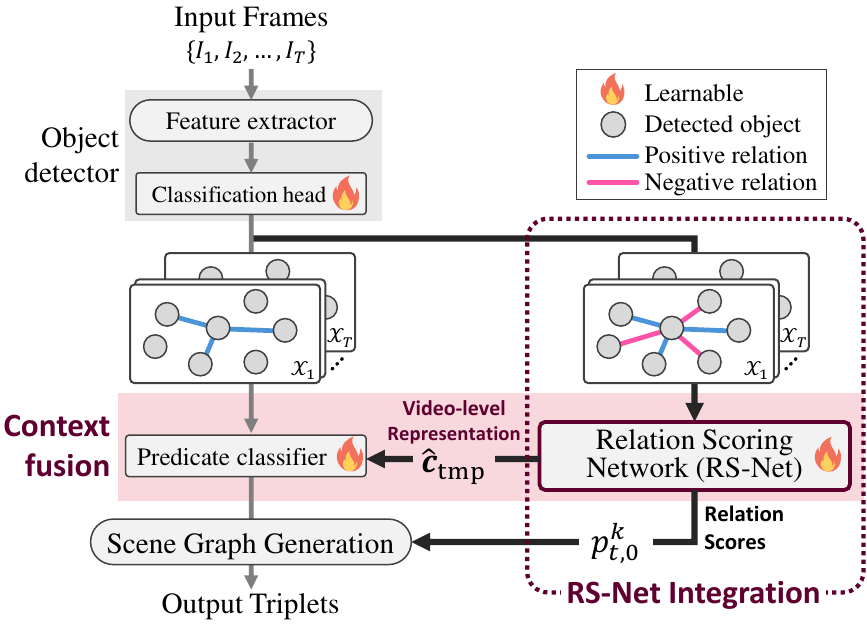}
    \caption{
        Integration of the RS-Net into existing DSGG frameworks for relation scoring. 
        Thanks to its modular design, RS-Net can be incorporated into various DSGG frameworks without requiring major structural modifications.
        }
    \label{fig:figure_3}
\end{figure}

\subsection{RS-Net Integration into Existing DSGG}

We integrate the proposed RS-Net into existing dynamic scene graph generation (DSGG) frameworks~\cite{feng2023exploiting, pu2023spatial} through a modular design that requires minimal architectural changes and supports end-to-end training.
Unlike prior methods that train predicate classifiers using only positive relations, RS-Net leverages all detected object pairs--including those without annotated relations--enabling the model to learn fine-grained relational importance across both positive and negative pairs.
Moreover, while conventional DSGG approaches rely on limited temporal windows (e.g., 2–3 frames), RS-Net provides entire video-level contextual reasoning.

As illustrated in Fig.~\ref{fig:figure_3}, the integration consists of two key steps:
(1) injecting the video-level temporal context token $\hat{\mathbf{c}}_{\mathrm{tmp}}$ into relation representations via a context fusion step, and
(2) incorporating RS-Net's relation scores $\mathbf{p}_t^k$ into the predicate classification process for each object pair.
By enriching relation representations with temporal context and providing explicit relation scores, RS-Net enhances performance of existing DSGG frameworks.

Context fusion refines the relation representation at the final stage before predicate classification by incorporating video-level temporal context.
Each relation feature $\mathbf{r}_t^k$ is first extracted as an intermediate representation from the predicate classification branch of the existing framework.
$\mathbf{r}_t^k$ is then concatenated with the video-level context token $\hat{\mathbf{c}}_{\mathrm{tmp}}$, enabling the model to encode both local spatial cues and global temporal semantics.
The fused representation is then processed by three linear heads, each specialized for a predicate type: attention, spatial, or contact.
Although the operation is simple in structure, it greatly improves how the model understands object relations.
Also, since it does not require any changes to the existing DSGG framework, our method can be easily applied to different baseline models.

During inference, the baseline model predicts relations for all detected object pairs.
In general, the triplet score for the $k$-th relation in frame $t$ in existing DSGG frameworks is computed as:
$s_{\text{sub}} \times s_{\text{obj}} \times s_{\text{rel}},$
where $s_{\text{sub}}$ and $s_{\text{obj}}$ denote the detection confidence scores for the subject and object, respectively, and
$s_{\text{rel}}$ is the confidence score from the predicate classifier.
With the integration of RS-Net, we further incorporate the relation score estimated by RS-Net to enhance the triplet score:
\begin{equation}
\text{Triplet score}_{t}^k = s_{\text{sub}} \times s_{\text{obj}} \times s_{\text{rel}} \times p_{t,0}^k,
\end{equation}
where $p_{t,0}^k$ denotes the relation score estimated by RS-Net, reflecting the contextual relevance of the relation.
This additional weighting helps the model suppress contextually irrelevant relations while emphasizing those that are semantically coherent, by leveraging both spatial and temporal context.

\subsection{Loss functions}
\label{sec:learning_RS-Net}
The proposed RS-Net, integrated into the baseline DSGG framework, jointly learns object detection, predicate classification, and  relation scoring for scene graph generation.
First, we use the cross-entropy loss to train the model to predict object categories, defined as:
\begin{equation}
\mathcal{L}_{od} = - \sum_{t,i} \mathbf{g}^i_t \log({\mathbf{d}}^i_t),
\end{equation}
where $\mathbf{g}^i_t$ is the one-hot encoded ground-truth label vector and $\mathbf{d}^i_t$ is the predicted category distribution for object $i$.
The predicate classification task reflects a multi-label setting, where multiple predicate categories may co-occur for a given object pair.
To this end, we formulate the predicate classification loss using the multi-label margin loss, defined as:
\begin{equation}
\mathcal{L}_{rel} = \sum_{m \in \mathcal{S}^+} \sum_{n \in \mathcal{S}^-} \max\left(0, 1 - s_m + s_n\right),
\end{equation}
where, $\mathcal{S}^+$ denotes the indices of the annotated predicates and $\mathcal{S}^-$ denotes the indices of the predicates not in the annotation.
$s_m$ and $s_n$ are the predicted confidence scores for the $m$-th and $n$-th predicates.

To optimize the proposed RS-Net when integrated into existing DSGG frameworks, we introduce a relation scoring loss that jointly trains the spatial-temporal context encoders and the relation scoring decoder. 
To focus learning on more challenging relation examples, we adopt the focal loss~\cite{lin2017focal} formulation:
\begin{equation}
\mathcal{L}_{RSN} = - \sum_{t,k} \sum_{c \in \{0,1\}} \alpha_c \, (1 - p_{t,c}^k)^{\gamma} \,  \log(p_{t,c}^k),
\end{equation}
where $\alpha_c$ is a balancing factor for $c$ (positive or negative relation), and $\gamma$ is a focusing parameter that reduces the loss contribution from well-classified relations.   
$p_{t,c}^k$ denotes the predicted probability for relation $\hat{\mathbf{x}}_t^k$ by relation scoring decoder. 
This formulation down-weights easy examples and emphasizes harder ones, helping the model better identify contextually meaningful relations.

Finally, the total loss is defined as:
\begin{equation}
\mathcal{L}_{total} = \mathcal{L}_{od} + \mathcal{L}_{rel} + \mathcal{L}_{RSN} .
\end{equation}

\section{Experimental Results}
\subsection{Datasets and Evaluation Metric}
We conducted experiments on the \texttt{Action} \texttt{Genome} dataset~\cite{ji2020action}, which is built upon the Charades dataset~\cite{sigurdsson2017asynchronous}.
It consists of 234,253 frames extracted from 9,848 video clips.
It includes annotations for 36 entity types and 26 predicate types, which are grouped into three categories: Attention, Spatial, and Contact.
The Attention category indicates whether a person is looking at an object; the Spatial category describes the relative positions between objects; and the Contact category represents various physical interactions between humans and objects.
We adopt two evaluation settings: Scene Graph Detection (SGDET) and Scene Graph Classification (SGCLS).
In SGDET, the model detects objects from images and predicts predicates for each object pair.
An object is considered correctly detected if its predicted bounding box has an Intersection-over-Union (IoU) of at least 0.5 with the ground truth.
In SGCLS, ground truth bounding boxes are provided, and the model is required to predict both object categories and pairwise predicates.
Following~\cite{zellers2018neural}, all evaluations are conducted under the ``With Constraint'' setting, which ensures that each subject–object pair is assigned at most one predicate.

All experiments are evaluated using the Recall@$k$ (R@$k$) metric, where $k \in \{10, 20, 50\}$.
In the experiments, we also report Precision@$k$ to assess the quality of the generated scene graphs.
While Recall measures how many ground-truth relations are retrieved, Precision@$k$ evaluates how many top-scoring predictions are actually relevant.
This reflects how effectively our RS-Net filters out contextually irrelevant relations based on the predicted relation scores.

\newcolumntype{W}{>{\centering\arraybackslash\hsize=6\hsize}X}
\newcolumntype{C}{>{\centering\arraybackslash\hsize=4.2\hsize}X} 
\newcolumntype{G}{>{\centering\arraybackslash\hsize=5\hsize}X} 


\newcolumntype{K}{>{\centering\arraybackslash\hsize=3.5\hsize}X} 
\newcolumntype{T}{>{\centering\arraybackslash}X}  

\begin{table}[t]
\renewcommand{\arraystretch}{1.0} 
\resizebox{\columnwidth}{!}{ 
\begin{tabular}{l@{\hskip 3pt}ccc@{\hskip 3pt}ccc}
\toprule
\multirow{2}{*}{Methods} & \multicolumn{3}{c@{\hskip 3pt}}{SGDET} & \multicolumn{3}{c}{SGCLS} \\
\cmidrule(lr){2-4} \cmidrule(lr){5-7}
& R@10 & R@20 & R@50 & R@10 & R@20 & R@50 \\
\hline
VRD \cite{lu2016visual} & 19.2 & 24.5 & 26.0 & 32.4 & 33.3 & 33.3 \\
MSDN \cite{li2017scene} & 24.1 & 32.4 & 34.5 & 43.9 & 45.1 & 45.1 \\
Motif Freq \cite{zellers2018neural} & 23.7 & 31.4 & 33.3 & 40.8 & 41.9 & 41.9 \\
VCTREE \cite{tang2019learning} & 24.4 & 32.6 & 34.7 & 44.1 & 45.3 & 45.3 \\
GPS-Net \cite{lin2020gps} & 24.7 & 33.1 & 35.1 & 45.3 & 46.5 & 46.5 \\
STTran$^{\dagger}$ \cite{cong2021spatial} & 25.1 & 33.9 & 36.`9 & 45.7 & 46.7 & 46.7 \\
STKET$^{\dagger}$ \cite{pu2023spatial} & 26.4 & 35.0 & \textbf{37.8} & 46.6 & 47.6 & 47.6 \\
TEMPURA\cite{nag2023unbiased} & 28.1 & 33.4 & 34.9 & 47.2 & 48.9 & 48.9 \\
DSGDetr$^{\dagger}$ \cite{feng2023exploiting} & \underline{30.3} & 34.8 & 36.1 & 49.9 & \underline{51.1} & 51.1 \\
TD2-Net \cite{lin2024td2} & 28.7 & - & \underline{37.1} & \textbf{51.1} & - & \textbf{52.1} \\
STABILE \cite{zhuang2025spatial} & 29.9 & 34.2 & 35.0 & 49.6 & 50.7 & 50.7 \\
\hline

Ours--STTran & 28.3 & 34.8 & 37.0 & 46.3 & 47.4 & 47.4 \\
Ours--STKET & 28.9 & \textbf{35.6} & \textbf{37.8} & 46.7 & 47.8 & 47.8 \\
Ours--DSGDetr & \textbf{30.5} & \underline{35.0} & 36.3 & \underline{50.5} & \textbf{51.7} & \underline{51.7} \\
\bottomrule
\end{tabular}
}
\caption{
        Performance comparison on the \texttt{Action} \texttt{Genome} dataset.
        Recall scores are reported for both SGDET and SGCLS settings.
        The best results are shown in bold, and the second-best are underlined.
        $^{\dagger}$ mark represents models retrained in our experimental environment.
        }
\label{tab:table_1}
\end{table}

\begin{table*}[t]
\centering
\small
\setlength{\tabcolsep}{4pt}
\begin{tabularx}{\textwidth}{l | T T | T T | T T | T T}
\hline
 Methods & Params (M) & FPS & R@10 & R@20 & P@10 & P@20 & mR@10 & mR@20 \\
\hline
STTran$^{\dagger}$ \cite{cong2021spatial} & 126.3 & 0.74 & 25.1 & 33.9 & 17.9 & 14.2 & 16.5 & 20.8 \\
Ours--STTran & 158.6 & 0.75 & \makebox[1pt][l]{\textbf{28.3}}\phantom{28.3}\rlap{{\scriptsize\tr{\textbf{\textit{ +3.2}}}}} & \makebox[1pt][l]{\textbf{34.8}}\phantom{34.8}\rlap{{\scriptsize\tr{\textbf{\textit{ +0.9}}}}} & \makebox[1pt][l]{\textbf{20.7}}\phantom{20.7}\rlap{{\scriptsize\tr{\textbf{\textit{ +2.8}}}}} & \makebox[1pt][l]{\textbf{14.9}}\phantom{14.9}\rlap{{\scriptsize\tr{\textbf{\textit{ +0.7}}}}} & \makebox[1pt][l]{\textbf{18.9}}\phantom{18.9}\rlap{{\scriptsize\tr{\textbf{\textit{ +2.4}}}}} & \makebox[1pt][l]{\textbf{23.2}}\phantom{23.2}\rlap{{\scriptsize\tr{\textbf{\textit{ +2.4}}}}} \\
\hline
STKET$^{\dagger}$ \cite{pu2023spatial} & 219.3 & 0.61 & 26.4 & 35.0 & 18.9 & 14.5 & 14.1 & 19.1 \\
Ours--STKET & 274.5 & 0.71 & \makebox[1pt][l]{\textbf{28.9}}\phantom{28.9}\rlap{{\scriptsize\tr{\textbf{\textit{ +2.5}}}}} & \makebox[1pt][l]{\textbf{35.6}}\phantom{35.6}\rlap{{\scriptsize\tr{\textbf{\textit{ +0.6}}}}} & \makebox[1pt][l]{\textbf{21.2}}\phantom{21.2}\rlap{{\scriptsize\tr{\textbf{\textit{ +2.3}}}}} & \makebox[1pt][l]{\textbf{15.2}}\phantom{15.2}\rlap{{\scriptsize\tr{\textbf{\textit{ +0.7}}}}} & \makebox[1pt][l]{\textbf{16.3}}\phantom{16.3}\rlap{{\scriptsize\tr{\textbf{\textit{ +2.2}}}}} & \makebox[1pt][l]{\textbf{20.8}}\phantom{20.8}\rlap{{\scriptsize\tr{\textbf{\textit{ +1.7}}}}} \\
\hline
DSGDetr$^{\dagger}$ \cite{feng2023exploiting} & 186.2 & 0.68 & 30.3 & 34.8 & 22.1 & 15.1 & 16.4 & 19.3 \\
Ours--DSGDetr & 232.4 & 0.70 & \makebox[1pt][l]{\textbf{30.5}}\phantom{30.5}\rlap{{\scriptsize\tr{\textbf{\textit{ +0.2}}}}} & \makebox[1pt][l]{\textbf{35.0}}\phantom{35.0}\rlap{{\scriptsize\tr{\textbf{\textit{ +0.2}}}}} & \makebox[1pt][l]{\textbf{22.2}}\phantom{22.2}\rlap{{\scriptsize\tr{\textbf{\textit{ +0.1}}}}} & \makebox[1pt][l]{\textbf{15.2}}\phantom{15.2}\rlap{{\scriptsize\tr{\textbf{\textit{ +0.1}}}}} & \makebox[1pt][l]{\textbf{17.6}}\phantom{17.6}\rlap{{\scriptsize\tr{\textbf{\textit{ +1.2}}}}} & \makebox[1pt][l]{\textbf{20.4}}\phantom{20.4}\rlap{{\scriptsize\tr{\textbf{\textit{ +1.1}}}}} \\
\hline
\end{tabularx}
\caption{Performance comparison with computational efficiency metrics in the SGDET task. RS-Net consistently improves both standard metrics and mean Recall across all baselines while maintaining competitive efficiency. Bold italic numbers indicate performance gains over the corresponding baselines. $^{\dagger}$ mark represents models retrained in our experimental environment.}
\label{tab:sgdet_results}
\end{table*}

\begin{table}[t]
\centering
\small
\setlength{\tabcolsep}{6pt}
\begin{tabularx}{\columnwidth}{l | c T T T}
\hline
Methods & R@10 & R@20 & P@10 & P@20 \\
\hline
STTran$^{\dagger}$ \cite{cong2021spatial} & 45.7 & 46.7 & 51.0 & 50.3 \\
Ours--STTran & \makebox[1pt][l]{\textbf{46.3}}\phantom{46.3}\rlap{{\scriptsize\tr{\textbf{\textit{ +0.6}}}}} & \makebox[1pt][l]{\textbf{47.4}}\phantom{47.4}\rlap{{\scriptsize\tr{\textbf{\textit{ +0.7}}}}} & \makebox[1pt][l]{\textbf{52.5}}\phantom{52.5}\rlap{{\scriptsize\tr{\textbf{\textit{ +1.5}}}}} & \makebox[1pt][l]{\textbf{51.7}}\phantom{51.7}\rlap{{\scriptsize\tr{\textbf{\textit{ +1.4}}}}} \\
\hline
STKET$^{\dagger}$ \cite{pu2023spatial} & 46.6 & 47.6 & 52.9 & 52.1 \\
Ours--STKET & \makebox[1pt][l]{\textbf{46.7}}\phantom{46.7}\rlap{{\scriptsize\tr{\textbf{\textit{ +0.1}}}}} & \makebox[1pt][l]{\textbf{47.8}}\phantom{47.8}\rlap{{\scriptsize\tr{\textbf{\textit{ +0.2}}}}} & \makebox[1pt][l]{\textbf{53.1}}\phantom{53.1}\rlap{{\scriptsize\tr{\textbf{\textit{ +0.2}}}}} & \makebox[1pt][l]{\textbf{52.3}}\phantom{52.3}\rlap{{\scriptsize\tr{\textbf{\textit{ +0.2}}}}} \\
\hline
DSGDetr$^{\dagger}$ \cite{feng2023exploiting} & 49.9 & 51.1 & 56.6 & 55.8 \\
Ours--DSGDetr & \makebox[1pt][l]{\textbf{50.5}}\phantom{50.5}\rlap{{\scriptsize\tr{\textbf{\textit{ +0.6}}}}} & \makebox[1pt][l]{\textbf{51.7}}\phantom{51.7}\rlap{{\scriptsize\tr{\textbf{\textit{ +0.6}}}}} & \makebox[1pt][l]{\textbf{57.0}}\phantom{57.0}\rlap{{\scriptsize\tr{\textbf{\textit{ +0.4}}}}} & \makebox[1pt][l]{\textbf{56.3}}\phantom{56.3}\rlap{{\scriptsize\tr{\textbf{\textit{ +0.5}}}}} \\
\hline
\end{tabularx}
\caption{Performance comparison in the SGCLS task. Bold italic numbers indicate performance gains over the corresponding baselines. $^{\dagger}$ mark represents models retrained in our experimental environment.}
\label{tab:sgcls_results}
\end{table}


\subsection{Implementation Details}
Following prior works~\cite{feng2023exploiting, pu2023spatial}, we adopt Faster R-CNN~\cite{ren2015faster} with a ResNet-101~\cite{he2016deep} backbone for object detection.
The detector is optimized using the AdamW optimizer~\cite{loshchilov2017decoupled}, with a learning rate of \(1 \times 10^{-5}\) and a batch size of one video per iteration.
For the RS-Net architecture, the MLP head consists of three layers with ReLU activation functions to generate relation scores. Our proposed Spatial Context Encoder and Temporal Context Encoder employ 2 and 4 transformer encoder blocks respectively, each with 4 attention heads. All models are trained for 10 epochs on two NVIDIA A100 GPUs.

A key challenge in training RS-Net lies in the inherent characteristics of video scene graph data. 
In video scenes, most object pairs merely coexist within the same frame without meaningful semantic associations, resulting in an overwhelming number of negative relation candidates. 
Moreover, many of these negative samples share the same object class combinations as positive relations, but exhibit poor spatial alignment (i.e., low IoU with ground truth).
This results in noisy training examples that can confuse the model.

To address this issue and improve training efficiency, we apply a negative sampling strategy.
Specifically, we exclude negative samples that share the same object class combinations as positive relations within the same frame, if their bounding boxes have an IoU greater than 0.5.
This strategy reduces training noise and helps the model focus on learning meaningful relation distinctions.
After this sampling step, if the number of negative relations still exceeds 1.2 times that of the positives, we randomly discard the excess negatives to maintain balanced training.

\subsection{Performance Comparison}
\label{sec:4.3}

We evaluated the performance of the proposed RS-Net by integrating it into three representative baselines: STTran~\cite{cong2021spatial}, DSG-DETR~\cite{feng2023exploiting}, and STKET~\cite{pu2023spatial} using the \texttt{Action} \texttt{Genome} dataset~\cite{ji2020action}.
As shown in Tab.~\ref{tab:table_1}, our method achieves state-of-the-art performance under the challenging SGDET setting, with Ours--DSG-DETR reaching 30.5\% R@10.
This improvement is particularly notable in the SGDET setting, demonstrating the effectiveness of RS-Net in capturing contextual relevance.
The other baselines, STTran and STKET, also showed substantial improvements in R@10, with gains of 3.2\% and 2.5\%, respectively.
This suggests that the proposed RS-Net effectively incorporates relation importance and video-level context information, regardless of the baseline frameworks.

In the SGCLS setting, our method improved the performance of its baseline, DSG-DETR, achieving 50.5\% R@10 and 51.7\% R@20, with gains of 0.6\% in both metrics.
These results demonstrate that the proposed RS-Net consistently yields meaningful performance improvements under both evaluation settings, achieving superior results across all metrics.
In addition, RS-Net is highly flexible and easy to integrate into existing frameworks, suggesting that even higher DSGG performance may be achieved when combined with stronger baseline architectures.

Table~\ref{tab:sgdet_results} compares the performance of three representative baseline models with and without RS-Net integration under the SGDET setting.
While prior works mainly focus on recall-based evaluation, we argue that precision is equally important in DSGG. 
It reflects the model’s ability to suppress irrelevant or noisy relations--an essential aspect of real-world DSGG performance.
Across all baseline frameworks, RS-Net leads to improvements in both recall and precision.
This result supports that explicitly modeling relation importance through contextual scoring is effective for enhancing the quality of scene graph generation in dynamic video settings.

In addition to recall and precision, we also report mean Recall, which provides a more balanced evaluation under the long-tailed distribution of relationships by equally weighting frequent and infrequent classes. RS-Net achieves consistent improvements in mR across all baselines, with gains of up to 2.4\% on STTran and comparable improvements on STKET and DSGDetr. These results highlight that our context-aware scoring mechanism not only enhances overall accuracy but also effectively mitigates the impact of class imbalance, demonstrating strong generalizability across different architectures.

It should be noted that although the RS-Net module adopts a uniform architectural design, the number of parameters differs slightly across baselines. Interestingly, while the integration of RS-Net inevitably increases the number of parameters, we observe a slight improvement in FPS. This somewhat unexpected yet consistent outcome aligns with prior observations that larger models or additional components can sometimes improve throughput by enabling more efficient GPU parallelism or aligning tensor dimensions with optimized kernel executions~\cite{dong2017more, kosaian2021boosting}. We hypothesize that RS-Net facilitates more favorable computational alignment and reduces certain runtime overheads, thereby compensating for the additional parameter cost.

Table~\ref{tab:sgcls_results} compares the performance of three representative baseline models with and without RS-Net integration under the SGCLS setting.
RS-Net again provides consistent improvements across all baselines, with particularly notable gains in precision. Since precision reflects the ability to suppress irrelevant or noisy relations, these results further confirm that our contextual scoring mechanism substantially improves the reliability of generated scene graphs.

\begin{figure*}[t]  
    \centering
    \includegraphics[width=\textwidth]{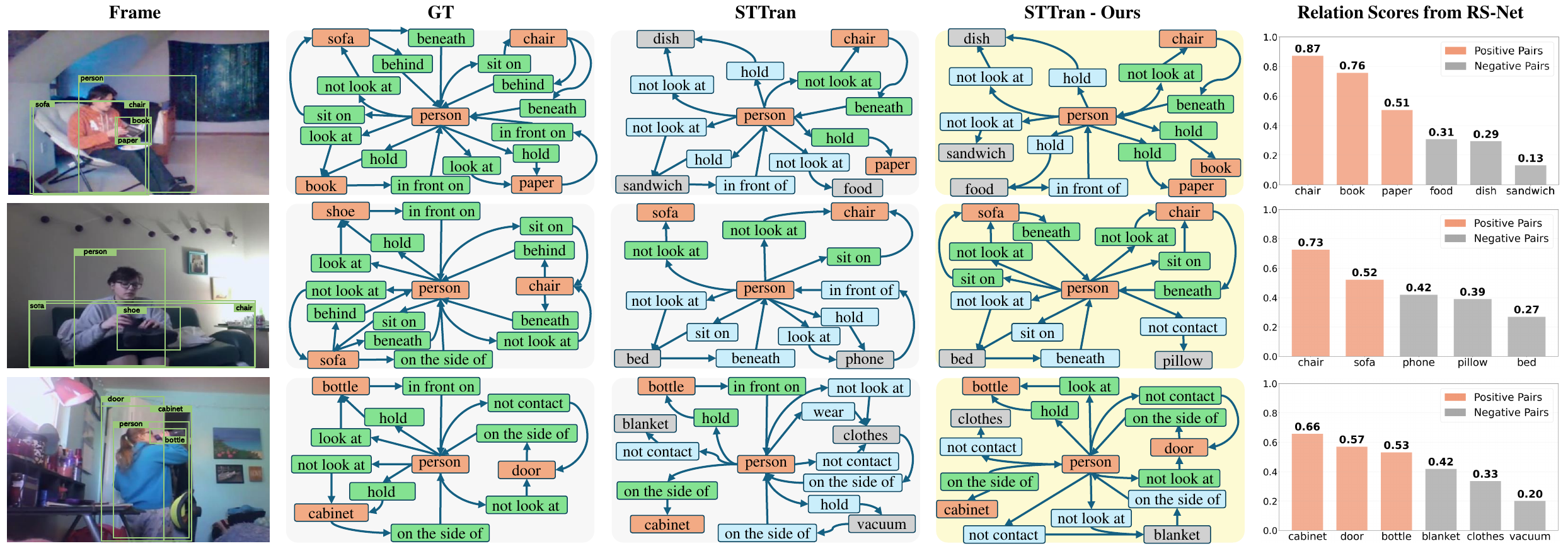}
    \caption{
        Qualitative comparisons between STTran and STTran with our RS-Net.
        The 3rd and 4th columns present the top predictions with highest confidence based on R@10 evaluation results in the SGDET setting.
        Orange and gray indicate detected objects that are involved in ground-truth relations and those that are not involved. 
        Green and blue indicate correctly and incorrectly predicted predicates, respectively.
        The 5th column shows the relation scores between person and objects predicted by RS-Net.
        }
    \label{fig:qualitative_1}
\end{figure*}

Figure~\ref{fig:qualitative_1} illustrates qualitative results comparing the original STTran with our RS-Net–enhanced version (Ours-STTran).
Compared to the baseline, applying RS-Net results in more correctly predicted relations (green boxes) being ranked higher.
During this process, detected object pairs with actual relational evidence in the input frame (orange boxes) are also re-ranked accordingly, contributing to more accurate scene graph construction.
The fifth column in Fig.~\ref{fig:qualitative_1} shows the relation scores $p^k_{t,0}$ predicted by RS-Net for the interaction between the person and the $k$-th object in frame $t$.
RS-Net assigns higher relation scores ($>0.51$) to object pairs that are contextually related in the video, while assigning lower scores ($<0.42$) to those without meaningful interactions.
The relation scores computed by RS-Net enable the re-ranking of triplets, elevating ground-truth relations to higher ranks.

\begin{figure}[t]
    \centering
    \includegraphics[width=1\linewidth]{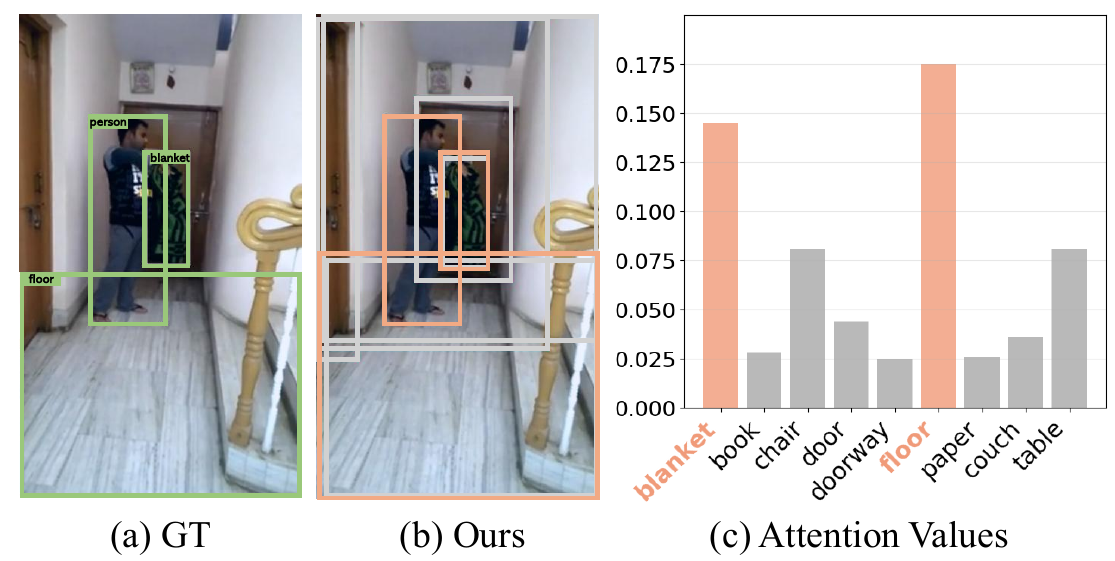}
    \caption{
         Attention scores visualization of the frame-level context token in the Spatial Context Encoder.
(a) Ground-truth subject and object instances, (b) detection results from our method, and (c) attention values from the frame-level context token to relation representations.
Orange color indicates object instances involved in ground-truth relations.
            }
    \label{fig:figure_5}
\end{figure}

\begin{table}[t]
\centering
\small
\begin{tabularx}{\columnwidth}{ l | c  c | c  c } \hline
Methods & R@10 & R@20 & P@10 & P@20 \\ \hline
No temporal encoder & 28.0 & 34.6 & 20.6 & 14.8 \\ 
Mean token $\mathbf{\hat{c}}_{\mathrm{mean}}$   & 28.1 & 34.6 & \textbf{20.7} & 14.8 \\ 
Learnable token $\mathbf{\hat{c}}_{\mathrm{tmp}}$ & \textbf{28.3} & \textbf{34.8} & \textbf{20.7} & \textbf{14.9} \\ \hline
No context fusion    & 28.0 & 34.7 & 20.6 & 14.8 \\
With context fusion  & \textbf{28.3} & \textbf{34.8} & \textbf{20.7} & \textbf{14.9} \\ \hline
\end{tabularx}
\caption{Ablation study. Effectiveness of proposed methods.}
\label{tab:ablation_1}
\end{table}

\subsection{Ablation Study on RS-Net Components}
We analyze the impact of each component in the proposed methods.
All ablation studies are conducted on the \texttt{Action} \texttt{Genome} dataset under the ``With Constraint'' setting in the SGDET task.
We use Ours--STTran as the base model for this experiment.

Table~\ref{tab:ablation_1} shows the effect of learning frame-level spatial context tokens $\hat{\mathbf{c}}_t$ and a video-level temporal context token $\mathbf{\hat{c}}_{\mathrm{tmp}}$.
For comparison, we conducted additional experiments: one excluding the temporal context encoder (No temporal encoder), and another using a temporal context token simply computed as the mean of frame-level spatial context tokens, defined as $\mathbf{\hat{c}}_{\mathrm{mean}} = \text{mean}([\mathbf{\hat{c}}_1, \mathbf{\hat{c}}_2, \ldots, \mathbf{\hat{c}}_T])$ (Mean token).
The learnable temporal context token yielded the best performance, followed by the mean-based token, while the model without any temporal context encoding showed the lowest performance.
These results highlight the importance of modeling video-level temporal information by aggregating spatial context tokens, which significantly contributes to improving DSGG performance.

As illustrated in Fig.\ref{fig:figure_3} and described in Sec.~{3.4}, we integrate RS-Net into existing frameworks by fusing the video-level context token $\mathbf{\hat{c}}_{\mathrm{tmp}}$ into their relation prediction process $\mathbf{r}^k_t$.
Table~\ref{tab:ablation_1} presents a comparison between models with this context fusion (With context fusion) and those without it (No context fusion).
As shown, incorporating context fusion consistently leads to higher DSGG performance.
Notably, integrating RS-Net requires no architectural changes or complex modifications to the baseline frameworks, yet enables efficient end-to-end training.

For further analysis, we visualize the attention values of the Spatial Context Encoder (Sec.~{3.3}), specifically illustrating how the frame-level context token $\mathbf{c}_t$ attends to intra-frame relations.
As shown in Fig.~\ref{fig:figure_5}, we observe that the attention values from the frame-level token are higher for object pairs involved in ground-truth relations (marked in olive).
This result suggests that the encoder learns to emphasize semantically relevant relations, guided by the contextual cues embedded in the frame-level token.

\section{Conclusions}

We presented RS-Net, a modular relation scoring network for dynamic scene graph generation.
RS-Net addresses two core limitations of existing DSGG methods: the lack of supervision for non-annotated object pairs and the inability to model long-range temporal context. To tackle these issues, RS-Net introduces a context-aware scoring mechanism that evaluates the semantic importance of object pairs by leveraging both spatial and video-level temporal representations. It integrates relation scores into a unified triplet scoring formulation, enhancing the model’s ability to identify meaningful relations while suppressing irrelevant ones.

Experiments on the \texttt{Action} \texttt{Genome} dataset show consistent improvements in Recall, Precision, and mean Recall, while maintaining competitive efficiency. Although the gains are relatively modest on strong baselines, advancing DSGG in such settings remains a challenging task and an important direction for future research.
Thanks to its lightweight and modular design, RS-Net can be seamlessly integrated into existing frameworks without architectural modifications, making it a practical and generalizable solution for real-world video understanding. These results further suggest that RS-Net can serve as a robust foundation for future DSGG research and diverse real-world video applications.


{
    \small
    \bibliographystyle{ieeenat_fullname}
    \bibliography{main}
}

\end{document}